\documentclass{article}

\usepackage[nonatbib,final]{nips_2018}

\usepackage[utf8]{inputenc}                 
\usepackage[T1]{fontenc}                    
\usepackage[hyperfootnotes=false]{hyperref} 
\usepackage{url}                            
\usepackage{booktabs}                       
\usepackage{amsfonts}                       
\usepackage{nicefrac}                       
\usepackage{microtype}                      
\usepackage{times}
\usepackage{epsfig}
\usepackage{pdfpages}
\usepackage{graphicx}
\usepackage{amsmath}
\usepackage{amssymb}
\usepackage{bm}
\usepackage{makecell}
\usepackage{multirow}
\usepackage{soul}
\usepackage[labelformat=simple]{subcaption}
\usepackage{lipsum}
\usepackage{wrapfig}

\newcommand\blfootnotee[1]{%
  \begingroup
  \renewcommand\thefootnote{}\footnote{*#1}%
  \addtocounter{footnote}{-1}%
  \endgroup
}

\newcommand{\eg}{e.g.}
\newcommand{\ie}{i.e.}
\newcommand{\etal}{et al.}
\newcommand{\etc}{etc.}

\begin{document}

\title{Deep Defense: Training DNNs with Improved Adversarial Robustness}

\author{Ziang Yan\textsuperscript{1*}\quad Yiwen Guo\textsuperscript{2,1*}\quad Changshui Zhang\textsuperscript{1}\\
 \textsuperscript{1}Institute for Artificial Intelligence, Tsinghua University (THUAI),\\
 State Key Lab of Intelligent Technologies and Systems,\\
 Beijing National Research Center for Information Science and Technology (BNRist),\\
 Department of Automation,Tsinghua University, Beijing, China\\
 \textsuperscript{2} Intel Labs China\\
 {\tt\small yza18@mails.tsinghua.edu.cn\quad yiwen.guo@intel.com\quad zcs@mail.tsinghua.edu.cn}
}

\maketitle

\begin{abstract}

 Despite the efficacy on a variety of computer vision tasks, deep neural networks (DNNs) are vulnerable to adversarial attacks, limiting their applications in security-critical systems.
 Recent works have shown the possibility of generating imperceptibly perturbed image inputs (a.k.a., adversarial examples) to fool well-trained DNN classifiers into making arbitrary predictions.
 To address this problem, we propose a training recipe named ``deep defense''.
 Our core idea is to integrate an adversarial perturbation-based regularizer into the classification objective, such that the obtained models learn to resist potential attacks, directly and precisely.
 The whole optimization problem is solved just like training a recursive network.
 Experimental results demonstrate that our method outperforms training with adversarial/Parseval regularizations by large margins on various datasets (including MNIST, CIFAR-10 and ImageNet) and different DNN architectures.
 Code and models for reproducing our results are available at \url{https://github.com/ZiangYan/deepdefense.pytorch}.
\end{abstract}

\section{Introduction}
Although deep neural networks (DNNs) have advanced the state-of-the-art of many challenging computer vision tasks, they are vulnerable to \emph{adversarial examples}~\cite{Szegedy2014} (\ie, generated images which seem perceptually similar to the real ones but are intentionally formed to fool learning models).\blfootnotee{The first two authors contributed equally to this work.}

A general way of synthesizing the adversarial examples is to apply worst-case perturbations to real images~\cite{Szegedy2014, Goodfellow2015, Moosavi2016, Carlini2017}.
With proper strategies, the required perturbations for fooling a DNN model can be 1000$\times$ smaller in magnitude when compared with the real images, making them imperceptible to human beings.
It has been reported that even the state-of-the-art DNN solutions have been fooled to misclassify such examples with high confidence~\cite{Kurakin2017}.
Worse, the adversarial perturbation can transfer across different images and network architectures~\cite{Moosavi2017}.
Such transferability also allows black-box attacks, which means the adversary may succeed without having any knowledge about the model architecture or parameters~\cite{Papernot2017}.

Though intriguing, such property of DNNs can lead to potential issues in real-world applications like self-driving cars and paying with your face systems.
Unlike certain instability against random noise, which is theoretically and practically guaranteed to be less critical~\cite{Fawzi2016, Szegedy2014}, the vulnerability to adversarial perturbations is still severe in deep learning.
Multiple attempts have been made to analyze and explain it so far~\cite{Szegedy2014, Goodfellow2015, Cisse2017, Hein2017}. 
For example, Goodfellow \etal~\cite{Goodfellow2015} argue that the main reason why DNNs are vulnerable is their linear nature instead of nonlinearity and overfitting.
Based on the explanation, they design an efficient $l_\infty$ induced perturbation and further propose to combine it with adversarial training~\cite{Szegedy2014} for regularization.
Recently, Cisse \etal~\cite{Cisse2017} investigate the Lipschitz constant of DNN-based classifiers and propose Parseval training. 
However, similar to some previous and contemporary methods, approximations to the theoretically optimal constraint are required in practice, making the method less effective to resist very strong attacks.

In this paper, we introduce ``deep defense'', an adversarial regularization method to train DNNs with improved robustness.
Unlike many existing and contemporaneous methods which make approximations and optimize possibly untight bounds, we precisely integrate a perturbation-based regularizer into the classification objective.
This endows DNN models with an ability of directly learning from attacks and further resisting them, in a principled way.
Specifically, we penalize the norm of adversarial perturbations, by encouraging relatively large values for the correctly classified samples and possibly small values for those misclassified ones.
As a regularizer, it is jointly optimized with the original learning objective and the whole problem is efficiently solved through being considered as training a recursive-flavoured network.
Extensive experiments on MNIST, CIFAR-10 and ImageNet show that our method significantly improves the robustness of different DNNs under advanced adversarial attacks, in the meanwhile {\bf no accuracy degradation is observed}.

The remainder of this paper is structured as follows.
First, we briefly introduce and discuss representative methods for conducting adversarial attacks and defenses in Section~\ref{sec:rel}.
Then we elaborate on the motivation and basic ideas of our method in Section~\ref{sec:dd}.
Section~\ref{sec:exp} provides implementation details of our method and experimentally compares it with the state-of-the-arts,
and finally, Section~\ref{sec:con} draws the conclusions.


\section{Related Work}\label{sec:rel}

\paragraph{Adversarial Attacks.} Starting from a common objective, many attack methods have been proposed.
Szegedy \etal~\cite{Szegedy2014} propose to generate adversarial perturbations by minimizing a vector norm using box-constrained L-BFGS optimization.
For better efficiency, Goodfellow \etal~\cite{Goodfellow2015} develop the fast gradient sign (FGS) attack, by choosing the sign of gradient as the direction of perturbation since it is approximately optimal under a $\ell_\infty$ constraint.
Later, Kurakin \etal~\cite{Kurakin2017} present an iterative version of the FGS attack by applying it multiple times with a small step size, and clipping pixel values on internal results.
Similarly, Moosavi-Dezfooli \etal~\cite{Moosavi2016} propose DeepFool as an iterative $l_p$ attack.
At each iteration, it linearizes the network and seeks the smallest perturbation to transform current images towards the linearized decision boundary.
Some more detailed explanations of DeepFool can be found in Section~\ref{subsec:ga}.
More recently, Carlini and Wagner~\cite{CW2017} reformulate attacks as optimization instances that can be solved using stochastic gradient descent to generate more sophisticated adversarial examples.
Based on the above methods, input- and network- agnostic adversarial examples can also be generated~\cite{Moosavi2017, Papernot2017}.



\paragraph{Defenses.} Resisting adversarial attacks is challenging.
It has been empirically studied that conventional regularization strategies such as dropout, weight decay and distorting training data (with random noise) do not really solve the problem~\cite{Goodfellow2015}.
Fine-tuning networks using adversarial examples, namely adversarial training~\cite{Szegedy2014}, is a simple yet effective approach to perform defense and relieve the problem~\cite{Goodfellow2015, Kurakin2017}, for which the examples can be generated either online~\cite{Goodfellow2015} or offline~\cite{Moosavi2016}.
Adversarial training works well on small datasets such as MNIST and CIFAR.
Nevertheless, as Kurakin \etal~\cite{Kurakin2017} have reported, it may result in a decreased benign-set accuracy on large-scale datasets like ImageNet.

An alternative way of defending such attacks is to train a detector, to detect and reject adversarial examples.
Metzen \etal~\cite{Metzen2017} utilize a binary classifier which takes intermediate representations as input for detection, and Lu \etal~\cite{Lu2017} propose to invoke an RBF-SVM operating on discrete codes from late stage ReLUs.
However, it is possible to perform attacks on the joint system if an adversary has access to the parameters of such a detector. 
Furthermore, it is still in doubt whether the adversarial examples are intrinsically different from the benign ones~\cite{Carlini2017}.

Another effective work is to exploit distillation~\cite{Papernot2016}, but it also slightly degrades the benign-set accuracy and may be broken by C\&W's attack~\cite{CW2017}.
Alemi \etal~\cite{Alemi2017} present an information theoretic method which helps on improving the resistance to adversarial attacks too.
Some recent and contemporaneous works also propose to utilize gradient masking~\cite{Papernot2018} as defenses~\cite{Dhillon2018, Xie2018, Buckman2018}.

Several regularization-based methods have also been proposed.
For example, Gu and Rigazio~\cite{Gu2015} propose to penalize the Frobenius norm of the Jacobian matrix in a layer-wise fashion.
Recently, Cisse \etal~\cite{Cisse2017} and Hein and Audriushchenko~\cite{Hein2017}  theoretically show that the sensitivity to adversarial examples can be controlled by the Lipschitz constant of DNNs and propose Parseval training and cross-Lipschitz regularization, respectively.
However, these methods usually require approximations, making them less effective to defend very strong and advanced adversarial attacks.  

As a regularization-based method, our Deep Defense is orthogonal to the adversarial training, defense distillation and detecting then rejecting methods.
It also differs from previous and contemporaneous regularization-based methods (\eg~\cite{Gu2015, Cisse2017, Hein2017, Ross2018}) in a way that it endows DNNs the ability of directly learning from adversarial examples and precisely resisting them.


\section{Our Deep Defense Method}\label{sec:dd}

\begin{figure*}[tp]
 \begin{center}
  \includegraphics[width=0.7\linewidth]{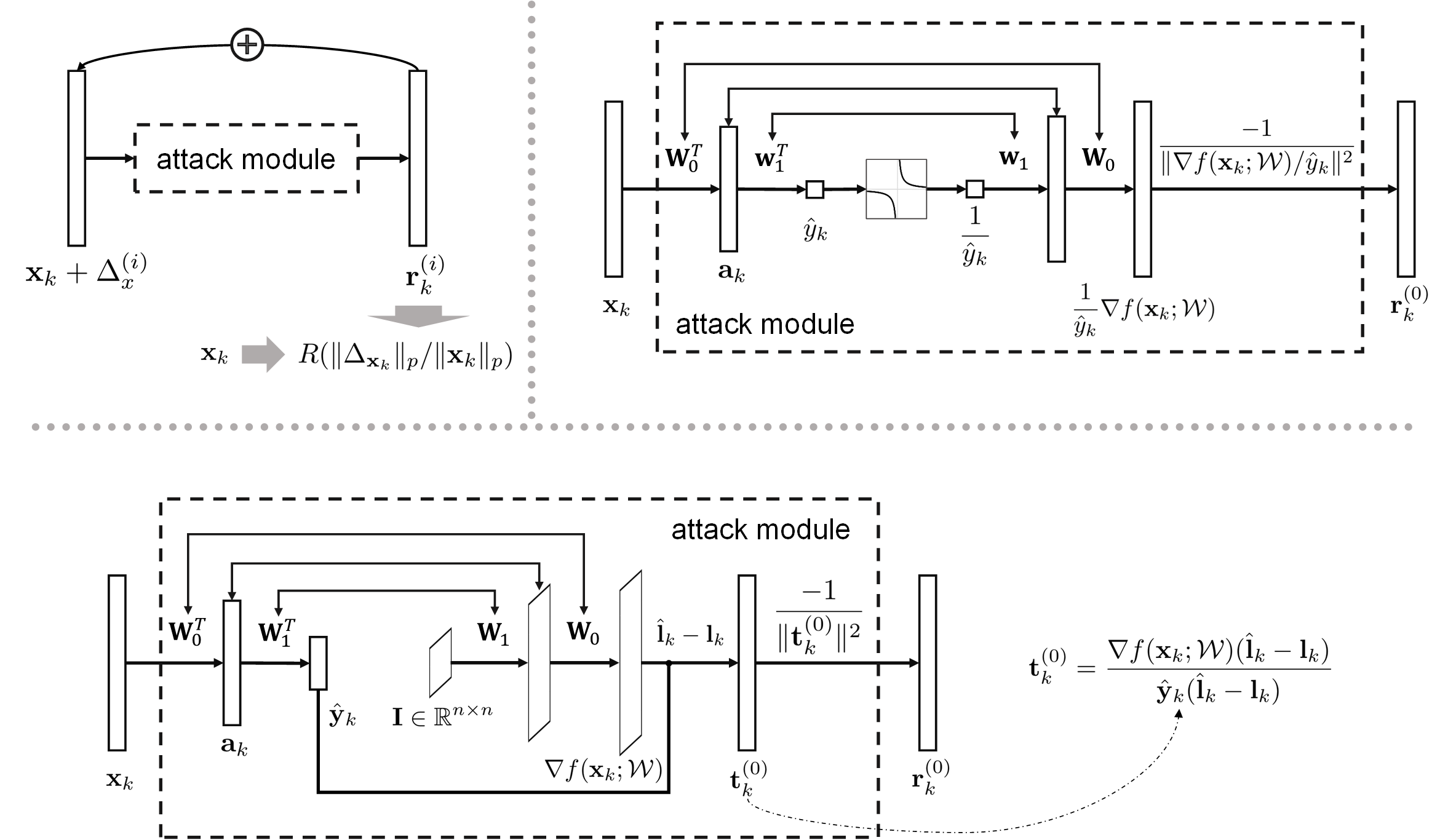}
 \end{center}
 \caption{Top left: The recursive-flavoured network which takes a reshaped image $\mathbf x_k$ as input and sequentially compute each perturbation component by using a pre-designed attack module. Top right: an example for generating the first component, in which the three elbow double-arrow connectors indicate weight-sharing fully-connected layers and index-sharing between ReLU activation layers. Bottom: the attack module for $n$-class ($n\geq 2$) scenarios.}
 \label{fig:1}
 \vskip -1em
\end{figure*}

Many methods regularize the learning objective of DNNs approximately, which may lead to a degraded prediction accuracy on the benign test sets or unsatisfactory robustness to advanced adversarial examples.
We reckon it can be more beneficial to incorporate advanced attack modules into the learning process and learn to maximize a margin.
In this section, we first briefly analyze a representative gradient-based attack and then introduce our solution to learn from it.

\subsection{Generate Adversarial Examples}\label{subsec:ga}

As discussed, a lot of efforts have been devoted to generating adversarial examples.
Let us take the $l_2$ DeepFool as an example here. It is able to conduct 100\% successful attacks on advanced networks.
Mathematically, starting from a binary classifier $f:\mathbb R^m \rightarrow \mathbb R$ which makes predictions (to the class label) based on the sign of its outputs, DeepFool generates the adversarial perturbation $\Delta_{\mathbf x}$ for an arbitrary input vector $\mathbf x \in \mathbb R^m$ in a heuristic way.
Concretely, $\Delta_{\mathbf x}= \mathbf r^{(0)}+...+\mathbf r^{(u-1)}$, in which the $i$-th ($0\leq i < u$) addend $\mathbf r^{(i)}$ is obtained by taking advantage of the Taylor's theorem and solving:
\begin{equation}\label{eq:1}
 \min_{\mathbf r} \| \mathbf r \|_2 \quad \mathrm{s.t.}\, f (\mathbf x + \Delta_\mathbf{x}^{(i)}) +  \nabla f (\mathbf x + \Delta_\mathbf{x}^{(i)})^T \mathbf r=0,
\end{equation}
in which $\Delta_\mathbf{x}^{(i)} := \sum _{j=0}^{i-1} \mathbf r^{(j)}$, function $\nabla f$ denotes the gradient of $f$ w.r.t. its input, and operator $\|\cdot\|_2$ denotes the $l_2$ (\ie, Euclidean) norm.
Obviously, Equation~(\ref{eq:1}) has a closed-form solution as:
\begin{equation}\label{eq:2}
 \mathbf r^{(i)} = -\frac{f (\mathbf x + \Delta_\mathbf{x}^{(i)})}{\|\nabla f(\mathbf x + \Delta_\mathbf{x}^{(i)})\|^2} \nabla f (\mathbf x + \Delta_x^{(i)}).
\end{equation}

By sequentially calculating all the $\mathbf r^{(i)}$s with~(\ref{eq:2}), DeepFool employs a faithful approximation to the $\Delta_{\mathbf x}$ of minimal $l_2$ norm.
In general, the approximation algorithm converges in a reasonably small number of iterations even when $f$ is a non-linear function represented by a very deep neural network, making it both effective and efficient in practical usage.
The for-loop for calculating $\mathbf r^{(i)}$s ends in advance if the attack goal $\mathrm{sgn}(f(\mathbf x + \Delta_\mathbf{x}^{(i)})) \neq \mathrm{sgn}(f(\mathbf x))$ is already reached at any iteration $i<u-1$.
Similarly, such strategy also works for the adversarial attacks to multi-class classifiers, which only additionally requires a specified target label in each iteration of the algorithm.

\subsection{Perturbation-based Regularization}\label{subsec:pr}

Our target is to improve the robustness of off-the-shelf networks without modifying their architectures, hence giving a $\|\Delta_{\mathbf x}\|_p$-based ($p\in [1,\infty$)) regularization to their original objective function seems to be a solution.

Considering the aforementioned attacks which utilize $\nabla f$ when generating the perturbation $\Delta_{\mathbf x}$~\cite{Szegedy2014, Goodfellow2015, Moosavi2016, Xie2017}, their strategy can be technically regarded as a function parameterized by the same set of learnable parameters as that of $f$.
Therefore, it is possible that we jointly optimize the original network objective and a scaled $\|\Delta_{\mathbf x} \|_p$ as a regularization for some chosen norm operator $\|\cdot\|_p$, provided $\|\Delta_{\mathbf x} \|_p$ is differentiable.
Specifically, given a set of training samples $\{(\mathbf x_k, \mathbf y_k)\}$ and a parameterized function $f$, we may want to optimize:
\begin{equation}\label{eq:3}
 \min_{\mathcal W}\, \sum_k L(\mathbf y_k, f(\mathbf x_k; \mathcal W)) + \lambda \sum_k R\left( -\frac{\|\Delta_{\mathbf x_k} \|_p}{\|\mathbf x_k\|_p}\right),
\end{equation}
in which the set $\mathcal W$ exhaustively collects learnable parameters of $f$, and $\|\mathbf x_k\|_p$ is a normalization factor for $\| \Delta_{\mathbf x_k} \|_p$.
As will be further detailed in Section~\ref{subsec:ra}, function $R$ should treat incorrectly and correctly classified samples differently, and it should be monotonically increasing on the latter such that it gives preference to those $f$s resisting small $\|\Delta_{\mathbf x_k} \|_p/\|\mathbf x_k\|_p$ anyway (\eg, $R(t)=\exp(t)$).
Regarding the DNN representations, $\mathcal W$ may comprise the weight and bias of network connections, means and variances of batch normalization layers~\cite{Ioffe2015}, and slops of the parameterized ReLU layers~\cite{He2015}.

\subsection{Network-based Formulation}

As previously discussed, we re-formulate the adversarial perturbation as $\Delta_{\mathbf x_k} = g(\mathbf x_k; \mathcal W)$, in which $g$ need to be differentiable except for maybe certain points, so that problem~(\ref{eq:3}) can be solved using stochastic gradient descent following the chain rule.
In order to make the computation more efficient and easily parallelized, an explicit formulation of $g$ or its gradient w.r.t $\mathcal W$ is required.
Here we accomplish this task by representing $g$ as a ``reverse'' network to the original one.
Taking a two-class multi-layer perceptron (MLP) as an example, we have $\mathcal W=\{\mathbf W_0, \mathbf b_0, \mathbf w_1, b_1\}$ and
\begin{equation}\label{eq:4}
 f(\mathbf x_k; \mathcal W) = \mathbf w_1^Th(\mathbf W_0^T\mathbf x_k + \mathbf b_0 ) + b_1,
\end{equation}
in which $h$ denotes the non-linear activation function and we choose $h(\mathbf W_0^T\mathbf x_k + \mathbf b_0):=\mathrm{max}(\mathbf W_0^T\mathbf x_k + \mathbf b_0, \mathbf 0)$ (\ie as the ReLU activation function) in this paper since it is commonly used.
Let us further denote $\mathbf a_k := h(\mathbf W_0^T\mathbf x_k + \mathbf b_0 )$ and $\hat y_k := f(\mathbf x_k; \mathcal W)$, then we have
\begin{equation}\label{eq:5}
 \nabla f(\mathbf x_k; \mathcal W) =\mathbf W_0 (\mathbf 1_{>0}(\mathbf a_k) \otimes \mathbf w_1),
\end{equation}
in which $\otimes$ indicates the element-wise product of two matrices, and $\mathbf 1_{>0}$ is an element-wise indicator function that compares the entries of its input with zero.

We choose $\Delta_{\mathbf x_k}$ as the previously introduced DeepFool perturbation for simplicity of notation~\footnote{Note that our method also naturally applies to some other gradient-based adversarial attacks.}.
Based on Equation~(\ref{eq:2}) and~(\ref{eq:5}), we construct a recursive-flavoured regularizer network (as illustrated in the top left of Figure~\ref{fig:1}) to calculate $R(-\| \Delta_{\mathbf x_k} \|_p / \|\mathbf x_k\|_p)$.
It takes image $\mathbf x_k$ as input and calculate each addend for $\Delta_{\mathbf x_k}$ by utilizing an incorporated multi-layer attack module (as illustrated in the top right of Figure~\ref{fig:1}).
Apparently, the original three-layer MLP followed by a multiplicative inverse operator makes up the first half of the attack module and its ``reverse'' followed by a norm-based rescaling operator makes up the second half.
It can be easily proved that the designed network is differentiable w.r.t. each element of $\mathcal W$, except for certain points.
As sketched in the bottom of Figure~\ref{fig:1}, such a network-based formulation can also be naturally generalized to regularize multi-class MLPs with more than one output neurons (\ie, $\hat{\mathbf y}_k\in \mathbb R^n$, $\nabla f(\mathbf x_k; \mathcal W) \in \mathbb R^{m\times n}$ and $n>1$).
We use $\mathbf I \in \mathbb R^{n\times n}$ to indicate the identity matrix, and $\hat{\mathbf l}_k$, $\mathbf l_k$ to indicate the one-hot encoding of current prediction label and a chosen label to fool in the first iteration, respectively.

Seeing that current winning DNNs are constructed as a stack of convolution, non-linear activation (\eg, ReLU, parameterized ReLU and sigmoid), normalization (\eg, local response normalization~\cite{Krizhevsky2012} and batch normalization), pooling and fully-connected layers, their $\nabla f$ functions, and thus the $g$ functions, should be differentiable almost everywhere.
Consequently, feasible ``reverse'' layers can always be made available to these popular layer types.
In addition to the above explored ones (\ie, ReLU and fully-connected layers), we also have deconvolution layers~\cite{Noh2015} which are reverse to the convolution layers, and unpooling layers~\cite{Zeiler2014} which are reverse to the pooling layers, \etc.
Just note that some learning parameters and variables like filter banks and pooling indices should be shared among them. 

\subsection{Robustness and Accuracy}\label{subsec:ra}

Problem~(\ref{eq:3}) integrates an adversarial perturbation-based regularization into the classification objective, which should endow parameterized models with the ability of learning from adversarial attacks and resisting them.
Additionally, it is also crucial not to diminish the inference accuracy on benign sets.
Goodfellow \etal~\cite{Goodfellow2015} have shown the possibility of fulfilling such expectation in a data augmentation manner.
Here we explore more on our robust regularization to ensure it does not degrade benign-set accuracies either.

Most attacks treat all the input samples equally~\cite{Szegedy2014, Goodfellow2015, Moosavi2016, Kurakin2017}, regardless of whether or not their predictions match the ground-truth labels.
It makes sense when we aim to fool the networks, but not when we leverage the attack module to supervise training.
Specifically, we might expect a decrease in $\|\Delta_{\mathbf x_k}\|_p/ \|\mathbf x_k\|_p$ from any misclassified sample $\mathbf x_k$, especially when the network is to be ``fooled'' to classify it as its ground-truth.
This seems different with the objective as formulated in~(\ref{eq:3}), which appears to enlarge the adversarial perturbations for all training samples.

Moreover, we found it difficult to seek reasonable trade-offs between robustness and accuracy, if $R$ is a linear function (\eg,~$R(z)=z$).
In that case, the regularization term is dominated by some extremely ``robust'' samples, so the training samples with relatively small $\|\Delta_{\mathbf x_k}\|_p/ \|\mathbf x_k\|_p$ are not fully optimized.
This phenomenon can impose a negative impact on the classification objective $L$ and thus the inference accuracy.
In fact, for those samples which are already ``robust'' enough, enlarging $\|\Delta_{\mathbf x_k}\|_p/ \|\mathbf x_k\|_p$ is not really necessary.
It is appropriate to penalize more on the currently correctly classified samples with abnormally small $\|\Delta_{\mathbf x_k} \|_p / \|\mathbf x_k\|_p$ values than those with relatively large ones (\ie, those already been considered ``robust'' in regard of $f$ and $\Delta_{\mathbf x_k}$).

To this end, we rewrite the second term in the objective function of Problem~(\ref{eq:3}) as
\begin{equation}\label{eq:6}
 \lambda\sum_{k\in \mathcal T} R\left(-c\frac{\|\Delta_{\mathbf x_k} \|_p}{\|\mathbf x_k\|_p}\right) +\lambda \sum_{k\in \mathcal F} R\left(d\frac{\|\Delta_{\mathbf x_k} \|_p}{\|\mathbf x_k\|_p}\right),
\end{equation}
in which $\mathcal F$ is the index set of misclassified training samples, $\mathcal T$ is its complement, $c, d>0$ are two scaling factors that balance the importance of different samples, and $R$ is chosen as the exponential function.
With extremely small or large $c$, our method treats all the samples the same in $\mathcal T$, otherwise those with abnormally small $\|\Delta_{\mathbf x_k} \|_p / \|\mathbf x_k\|_p$ will be penalized more than the others.

\begin{table}[th]
 \vspace{-1em}
 \caption{Test set performance of different defense methods. Column 4: prediction accuracies on benign examples. Column 5: $\rho_2$ values under the DeepFool attack. Column 6-8: prediction accuracies on the FGS adversarial examples. }\label{tab:rhoacc}
 \begin{center}\resizebox{0.85\linewidth}{!}{
   \begin{tabular}{cccccccc}
    \toprule
    Dataset                   & Network                  & Method       & Acc.          & $\rho_2$                               & Acc.@$0.2\epsilon_{\rm ref}$ & Acc.@$0.5\epsilon_{\rm ref}$ & Acc.@$1.0\epsilon_{\rm ref}$ \\
    \midrule
    \multirow{8}{*}{MNIST}    & \multirow{4}{*}{MLP}     & Reference    & 98.31\%       & 1.11$\times \mathrm{10}^{\mathrm{-1}}$ & 72.76\%                      & 29.08\%                      & 3.31\%                       \\
                              &                          & Par. Train   & 98.32\%       & 1.11$\times10^{-1}$                    & 77.44\%                      & 28.95\%                      & 2.96\%                       \\
                              &                          & Adv. Train I & 98.49\%       & 1.62$\times10^{-1}$                    & 87.70\%                      & 59.69\%                      & 22.55\%                      \\
                              &                          & Ours         & {\bf 98.65\%} & \bf 2.25$\bm{\times10^{-1}}$           & {\bf 95.04\%}                & {\bf 88.93\%}                & {\bf 50.00\%}                \\ \cmidrule(r){2-8}
                              & \multirow{4}{*}{LeNet}   & Reference    & 99.02\%       & 2.05$\times10^{-1}$                    & 90.95\%                      & 53.88\%                      & 19.75\%                      \\
                              &                          & Par. Train   & 99.10\%       & 2.03$\times10^{-1}$                    & 91.68\%                      & 66.48\%                      & 19.64\%                      \\
                              &                          & Adv. Train I & 99.18\%       & 2.63$\times10^{-1}$                    & 95.20\%                      & 74.82\%                      & 41.40\%                      \\
                              &                          & Ours         & {\bf 99.34\%} & \bf 2.84$\bm{\times10^{-1}}$           & {\bf 96.51\%}                & {\bf 88.93\%}                & {\bf 50.00\%}                \\
    \midrule
    \multirow{8}{*}{CIFAR-10} & \multirow{4}{*}{ConvNet} & Reference    & 79.74\%       & 2.59$\times10^{-2}$                    & 61.62\%                      & 37.84\%                      & 23.85\%                      \\
                              &                          & Par. Train   & 80.48\%       & 3.42$\times10^{-2}$                    & 69.19\%                      & 50.43\%                      & 22.13\%                      \\
                              &                          & Adv. Train I & 80.65\%       & 3.05$\times10^{-2}$                    & 65.16\%                      & 45.03\%                      & 35.53\%                      \\
                              &                          & Ours         & {\bf 81.70\%} & \bf 5.32$\bm{\times10^{-2}}$           & {\bf 72.15\%}                & {\bf 59.02\%}                & {\bf 50.00\%}                \\ \cmidrule(r){2-8}
                              & \multirow{4}{*}{NIN}     & Reference    & 89.64\%       & 4.20$\times10^{-2}$                    & 75.61\%                      & 49.22\%                      & 33.56\%                      \\
                              &                          & Par. Train   & 88.20\%       & 4.33$\times10^{-2}$                    & 75.39\%                      & 49.75\%                      & 17.74\%                      \\
                              &                          & Adv. Train I & 89.87\%       & 5.25$\times10^{-2}$                    & 78.87\%                      & 58.85\%                      & 45.90\%                      \\
                              &                          & Ours         & {\bf 89.96\%} & \bf 5.58$\bf\times10^{-2}$             & {\bf 80.70\%}                & {\bf 70.73\%}                & {\bf 50.00\%}                \\ \midrule
    \multirow{4}{*}{ImageNet} & \multirow{2}{*}{AlexNet} & Reference    & 56.91\%       & 2.98$\times\mathrm{10}^{\mathrm{-3}}$  & 54.62\%                      & 51.39\%                      & 46.05\%                      \\
                              &                          & Ours         & {\bf 57.11\%} & \bf 4.54$\bm{\times10^{-3}}$           & \bf 55.79\%                  & \bf 53.50\%                  & \bf 50.00\%                  \\ \cmidrule(r){2-8}
                              & \multirow{2}{*}{ResNet}  & Reference    & 69.64\%       & 1.63$\times\mathrm{10}^{\mathrm{-3}}$  & 63.39\%                      & 54.45\%                      & 41.70\%                      \\
                              &                          & Ours         & {\bf 69.66\%} & \bf 2.43$\bm{\times10^{-3}}$           & \bf 65.53\%                  & \bf 59.46\%                  & \bf 50.00\%                  \\ \bottomrule
   \end{tabular}}
 \end{center}
 \vspace{-1.8em}
\end{table}

\section{Experimental Results}\label{sec:exp}

In this section, we evaluate the efficacy of our method on three different datasets: MNIST, CIFAR-10 and ImageNet~\cite{Russakovsky2015}.
We compare our method with adversarial training and Parseval training (also known as Parseval networks).
Similar to previous works~\cite{Moosavi2016, Alemi2017}, we choose to fine-tune from pre-trained models instead of training from scratch.
Fine-tuning hyper-parameters can be found in the supplementary materials.
All our experiments are conducted on an NVIDIA GTX 1080 GPU.  
Our main results are summarized in Table~\ref{tab:rhoacc}, where the fourth column demonstrates the inference accuracy of different models on benign test images, the fifth column compares the robustness of different models to DeepFool adversarial examples, and the subsequent columns compare the robustness to FGS adversarial examples.
The evaluation metrics will be carefully explained in Section~\ref{subsec:em}.
Some implementation details of the compared methods are shown as follows.

\paragraph{Deep Defense.} There are three hyper-parameters in our method: $\lambda$, $c$ and $d$.
As previously explained in Section~\ref{subsec:ra}, they balance the importance of the model robustness and benign-set accuracy.
We fix $\lambda=15, c=25, d=5$ for MNIST and CIFAR-10 major experiments (except for NIN, $c=70$), and uniformly set $\lambda=5, c=500, d=5$ for all ImageNet experiments.
Practical impact of varying these hyper-parameters will be discussed in Section~\ref{subsec:expm}.
The Euclidean norm is simply chosen for $\|\cdot\|_p$.

\paragraph{Adversarial Training.} There exist many different versions of adversarial training~\cite{Szegedy2014, Goodfellow2015, Moosavi2016, Kurakin2017, Miyato2017, Madry2018}, partly because it can be combined with different attacks.
Here we choose two of them, in accordance with the adversarial attacks to be tested,  and try out to reach their optimal performance.
First we evaluate the one introduced in the DeepFool paper~\cite{Moosavi2016}, which utilizes a fixed adversarial training set generated by DeepFool, and summarize its performance in Table~\ref{tab:rhoacc} (see ``Adv. Train I'').
We also test Goodfellow et al.'s adversarial training objective~\cite{Goodfellow2015} (referred to as ``Adv. Train II'') and compare it with our method intensively (see supplementary materials), considering there exists trade-offs between accuracies on benign and adversarial examples.
In particular, a combined method is also evaluated to testify our previous claim of orthogonality.

\paragraph{Parseval Training.} Parseval training~\cite{Cisse2017} improves the robustness of a DNN by controlling its global Lipschitz constant.
Practically, a projection update is performed after each stochastic gradient descent iteration to ensure all weight matrices' Parseval tightness.
Following the original paper, we uniformly sample $30\%$ of columns to perform this update.
We set the hyper-parameter $\beta=0.0001$ for MNIST, and $\beta=0.0003$ for CIFAR-10 after doing grid search.

\subsection{Evaluation Metrics}\label{subsec:em}

This subsection explains some evaluation metrics adopted in our experiments.
Different $l_p$  (e.g., $l_2$ and $l_\infty$) norms have been used to perform attacks.
Here we conduct the famous FGS and DeepFool as representatives of $l_\infty$ and $l_2$ attacks and compare the robustness of obtained models using different defense methods.
As suggested in the paper \cite{Moosavi2016}, we evaluate model robustness by calculating
\begin{equation}\label{eq:7}
 \rho_2:=\frac{1}{|\mathcal{D}|}\sum_{k\in \mathcal D}\frac{\| \Delta_{\mathbf x_k}\|_2}{\| \mathbf x_k \|_2},
\end{equation}
in which $\mathcal{D}$ is the test set (for ImageNet we use its validation set), when DeepFool is used.

It is popular to evaluate the accuracy on a perturbed $\mathcal D$ as a metric for the FGS attack~\cite{Gu2015, Goodfellow2015, Cisse2017}.
Likewise, we calculate the smallest $\epsilon$ such that 50\% of the perturbed images are misclassified by our regularized models and denote it as $\epsilon_{\rm ref}$, then test prediction accuracies of those models produced by adversarial and Parseval training at this level of perturbation (abbreviated as ``Acc.@$1.0\epsilon_{\rm ref}$'' in Table~\ref{tab:rhoacc}).
Accuracies at lower levels of perturbations (a half and one fifth of $\epsilon_{\rm ref}$) are also reported.

Many other metrics will be introduced and used for further comparisons in supplementary materials.

\begin{figure}[t]
 \centering
 \vspace{-1em}
 \begin{subfigure}[b]{0.23\linewidth}
  \includegraphics[width=\linewidth]{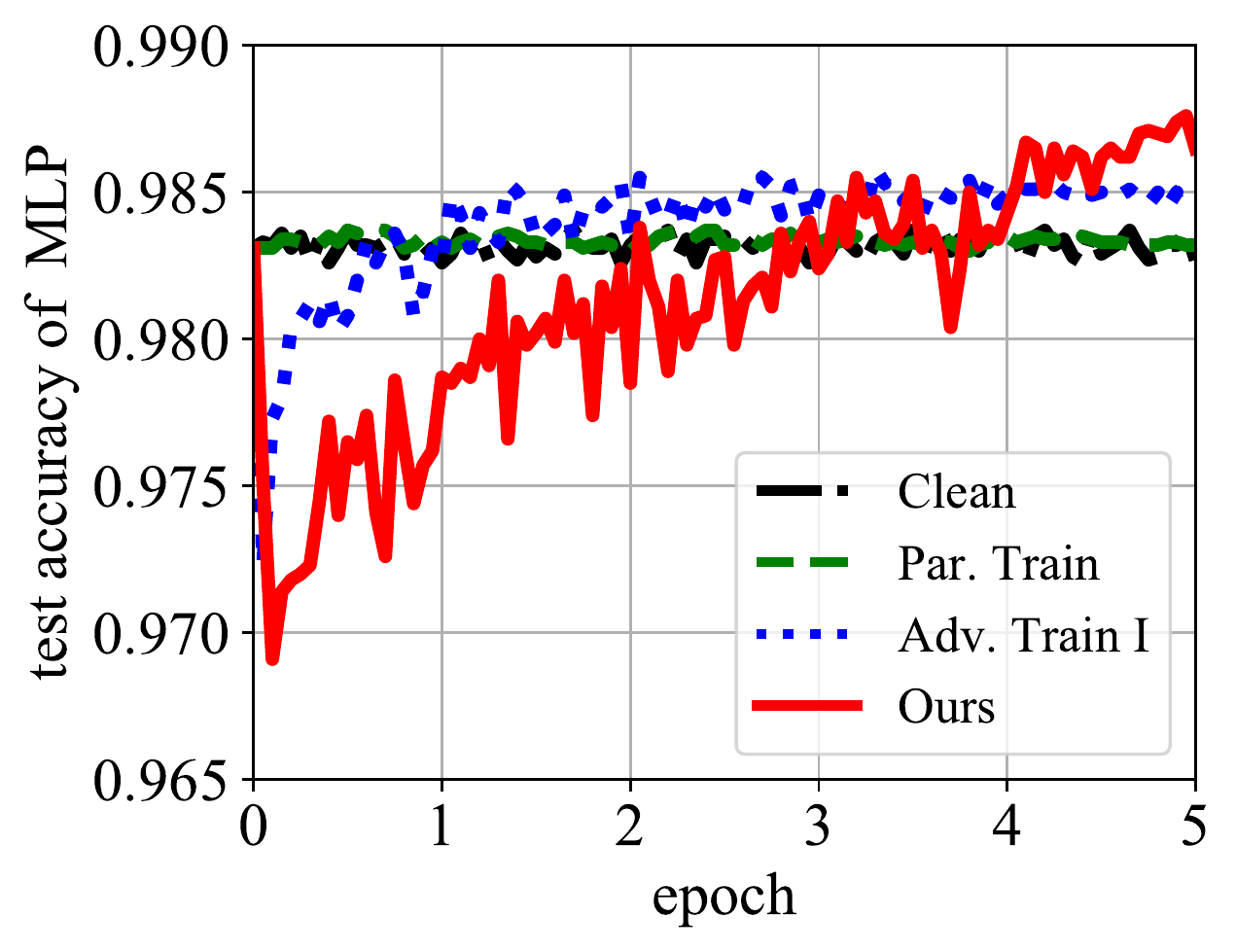}
 \end{subfigure}
 \begin{subfigure}[b]{0.23\linewidth}
  \includegraphics[width=\linewidth]{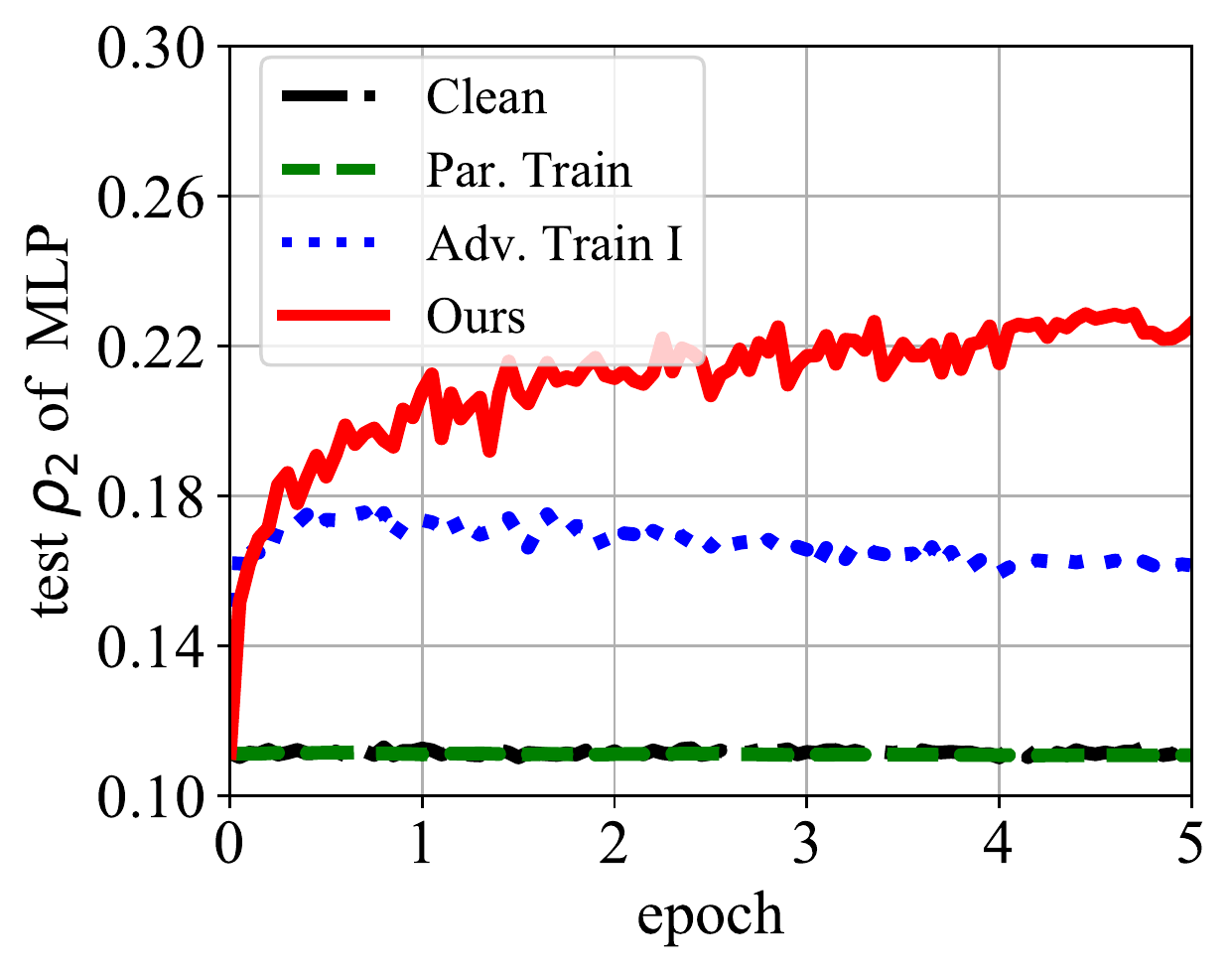}
 \end{subfigure}
 \begin{subfigure}[b]{0.23\linewidth}
  \includegraphics[width=\linewidth]{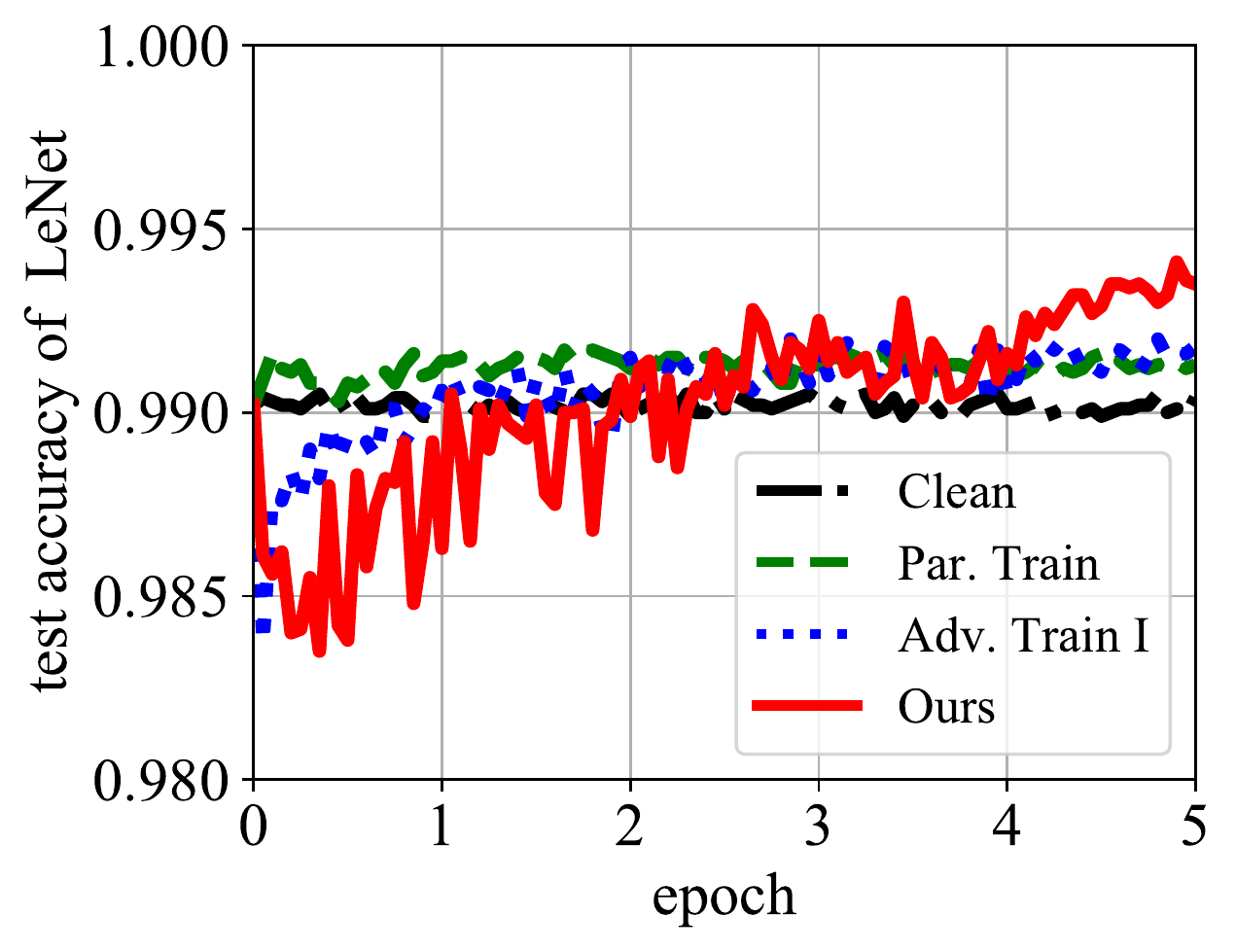}
 \end{subfigure}
 \begin{subfigure}[b]{0.23\linewidth}
  \includegraphics[width=\linewidth]{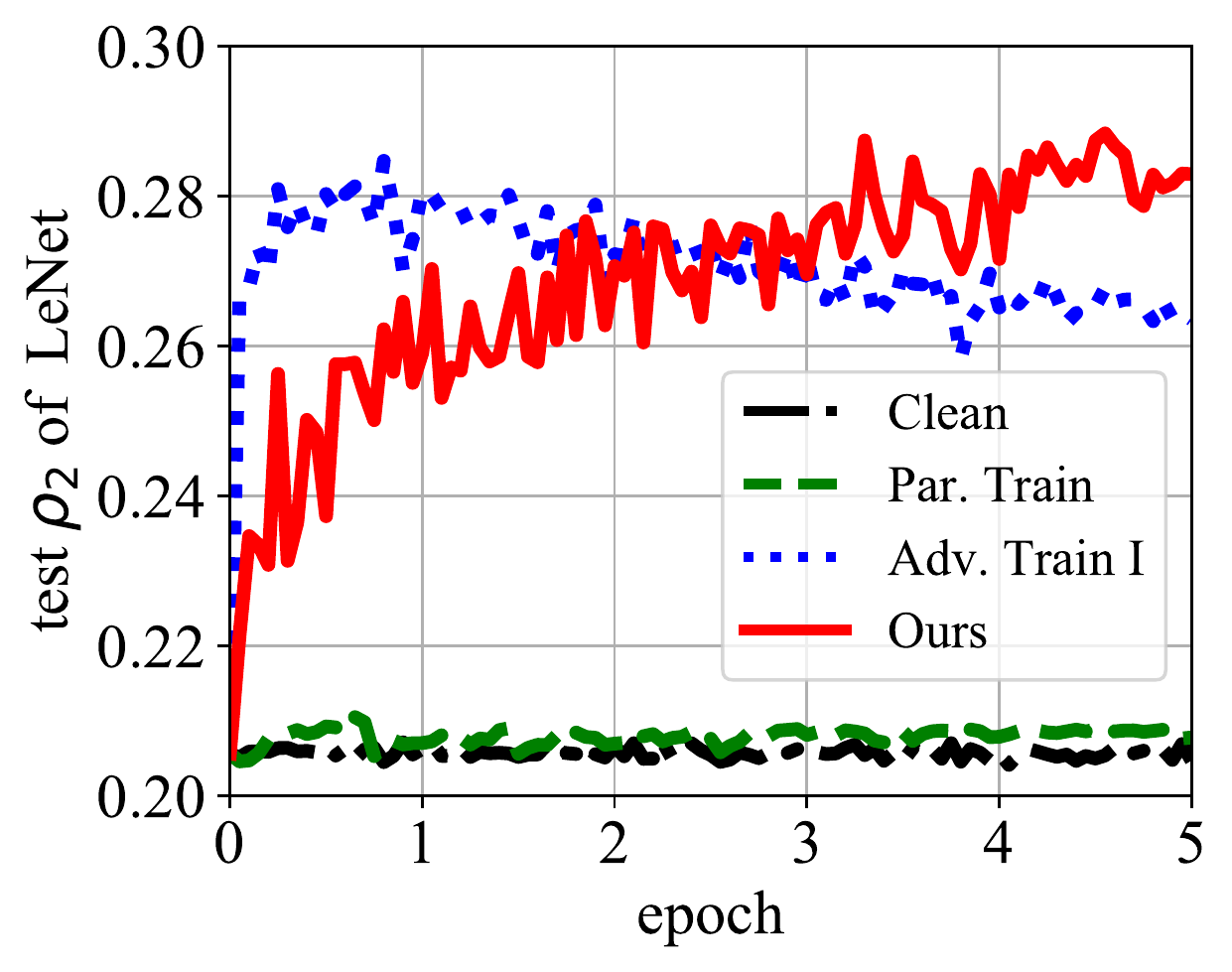}
 \end{subfigure}

 \caption{Convergence curves. From left to right: test accuracy and $\rho_2$ of MLP, and test accuracy and $\rho_2$ of LeNet. ``Clean'' indicates fine-tuning on benign examples. Best viewed in color.}\label{fig:mnistcifarcurvs}\vskip -0.2in
\end{figure}

\subsection{Exploratory Experiments on MNIST}\label{subsec:expm}

As a popular dataset for conducting adversarial attacks \cite{Szegedy2014, Goodfellow2015, Moosavi2016}, MNIST is a reasonable choice for us to get started.
It consists of 70,000 grayscale images, in which 60,000 of them are used for training and the remaining are used for test.
We train a four-layer MLP and download a LeNet~\cite{Lecun1999} structured CNN model~\footnote{\url{https://github.com/LTS4/DeepFool/blob/master/MATLAB/resources/net.mat}} as references (see supplementary materials for more details).
For fair comparisons, we use identical fine-tuning policies and hyper-parameters for different defense methods 
We cut the learning rate by 2$\times$ after four epochs of training because it can be beneficial for convergence.

\paragraph{Robustness and accuracy.}
The accuracy of different models (on the benign test sets) can be found in the fourth column of Table \ref{tab:rhoacc} and the robustness performance is compared in the last four columns.
We see Deep Defense consistently and significantly outperforms competitive methods in the sense of both robustness and accuracy, even though our implementation of Adv. Train I achieves slightly better results than those reported in~\cite{Moosavi2016}.
Using our method, we obtain an MLP model with {\bf over} $\bm{2\times}$ better robustness to DeepFool and an absolute error decrease of  46.69\% under the FGS attack considering $\epsilon=1.0\epsilon_\mathrm{ref}$, while the inference accuracy also increases a lot (from 98.31\% to {\bf 98.65\%} in comparison with the reference model.
The second best is Adv. Train I, which achieves roughly 1.5$\times$ and an absolute 19.24\% improvement under the DeepFool and FGS attacks, respectively.
Parseval training also yields models with improved robustness to the FGS attack, but they are still vulnerable to the DeepFool.
The superiority of our method holds on LeNet, and the benign-set accuracy increases from 99.02\% to {\bf 99.34\%} with the help of our method.

\begin{wrapfigure}{r}{0.4\textwidth}
 \centering
 \vspace{-1.5em}
 \includegraphics[width=1.0\linewidth, trim=0.1cm 0 0.5cm 2cm, clip]{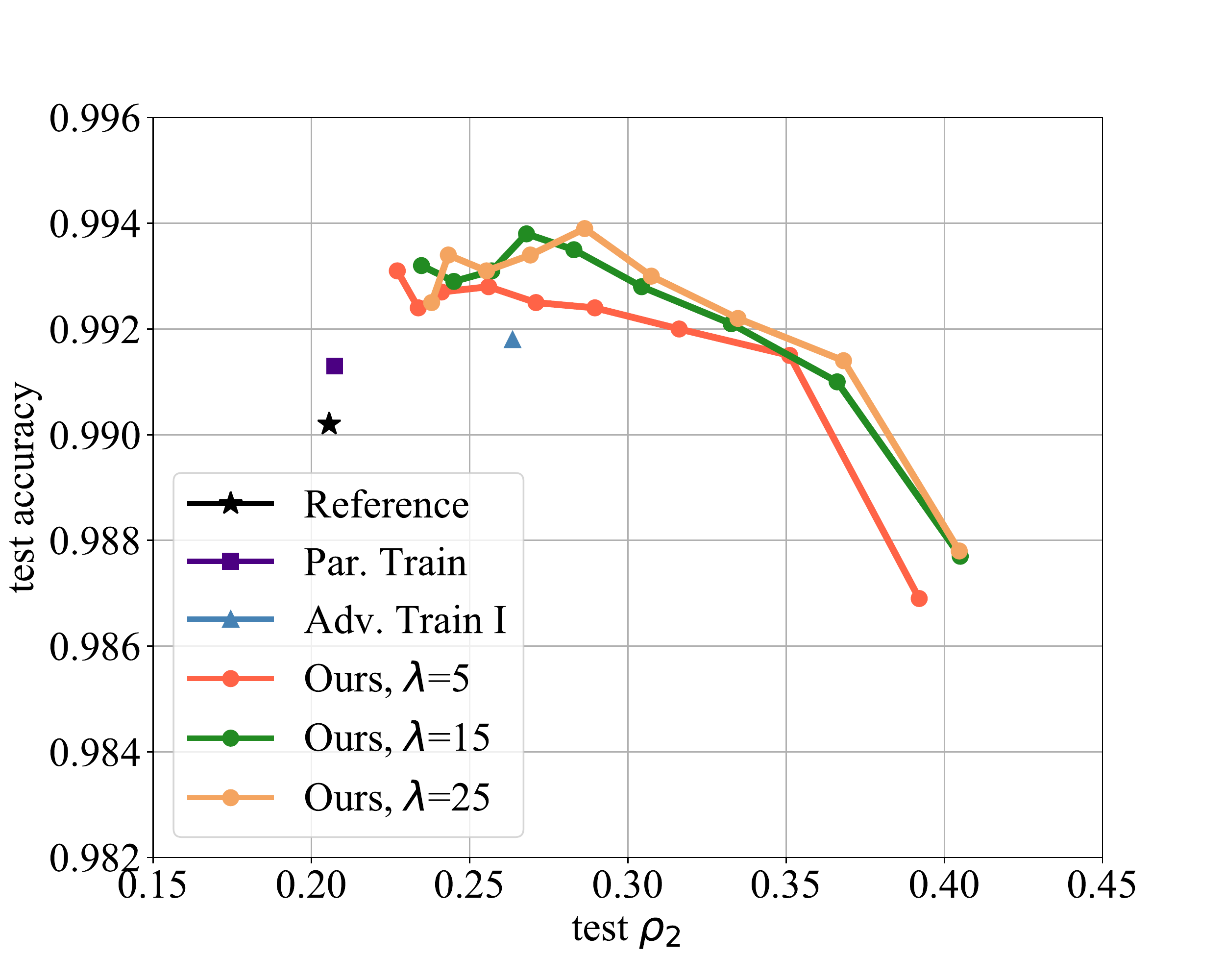}
 \caption{The performance of Deep Defense with varying hyper-parameters on LeNet. Best viewed in color.}
 \vspace{-1.5em}
 \label{fig:varycd}
\end{wrapfigure}

Convergence curves of different methods are provided in Figure~\ref{fig:mnistcifarcurvs}, in which the ``Clean'' curve indicates fine-tuning on the benign training set with the original learning objective.
Our method optimizes more sophisticated objective than the other methods so it takes longer to finally converge.
However, both its robustness and accuracy performance surpasses that of the reference models in only three epochs and keeps growing in the last two.
Consistent with results reported in~\cite{Moosavi2016}, we also observe growing accuracy and decreasing $\rho_2$ on Adv. Train I.

In fact, the benefit of our method to test-set accuracy for benign examples is unsurprising.
From a geometrical point of view, an accurate estimation of the optimal perturbation like our $\Delta_{\mathbf x_k}$ represents the distance from a benign example $\mathbf x_k$ to the decision boundary, so maximizing $\|\Delta_{\mathbf x_k}\|$ approximately maximizes the margin.
According to some previous theoretical works~\cite{Xu2012, Sokolic2017}, such regularization to the margin should relieve the overfitting problem of complex learning models (including DNNs) and thus lead to better test-set performance on benign examples.

\paragraph{Varying Hyper-parameters.} Figure~\ref{fig:varycd} illustrates the impact of the hyper-parameters in our method.
We fix $d=5$ and try to vary $c$ and $\lambda$ in $\{5, 10, 15, 20, 25, 30, 35, 40, 45\}$ and $\{5, 15, 25\}$, respectively.
Note that $d$ is fixed here because it has relatively minor effect on our fine-tuning process on MNIST.
In the figure, different solid circles on the same curve indicate different values of $c$.
From left to right, they are calculated with decreasing $c$, which means a larger $c$ encourages achieving a better accuracy but lower robustness.
Conversely, setting a very small $c$ (\eg, $c=5$) can yield models with high robustness but low accuracies.
By adjusting $\lambda$, one changes the numerical range of the regularizer.
A larger $\lambda$ makes the regularizer contributes more to the whole objective function.

\begin{wrapfigure}{r}{0.4\textwidth}
 \centering
 \vspace{-1em}
 \includegraphics[width=1.0\linewidth, trim=0.1cm 0 0.5cm 2cm, clip]{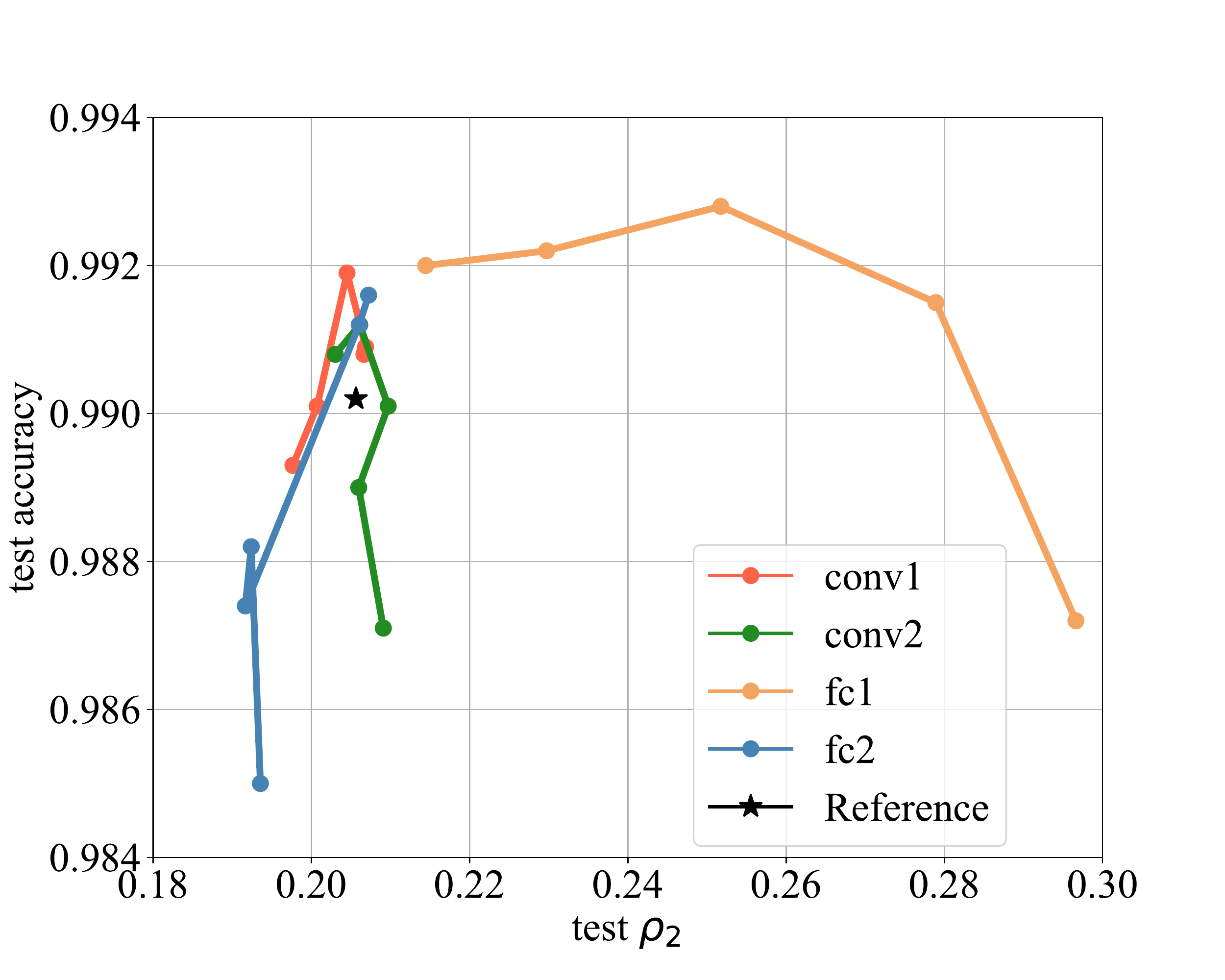}
 \caption{The performance of Deep Defense when only one layer is regularized for LeNet. Best viewed in color.}\label{fig:mask_grad}
\end{wrapfigure}

\vspace{-.8em}
\paragraph{Layer-wise Regularization.} We also investigate the importance of different layers to the robustness of LeNet with our Deep Defense method.
Specifically, we mask the gradient (by setting its elements to zero) of our adversarial regularizer w.r.t. the learning parameters (\eg, weights and biases) of all layers except one.
By fixing $\lambda=15$, $d=5$ and varying $c$ in the set $\{5, 15, 25, 35, 45\}$, we obtain 20 different models.
Figure~\ref{fig:mask_grad} demonstrates the $\rho_2$ values and benign-set accuracies of these models.
Different points on the same curve correspond to fine-tuning with different values of $c$ (decreasing from left to right).
Legends indicate the gradient of which layer is not masked.
Apparently, when only one layer is exploited to regularize the classification objective, optimizing ``fc1'' achieves the best performance.
This is consistent with previous results that ``fc1'' is the most ``redundant'' layer of LeNet~\cite{Han2015, Guo2016}. %

\subsection{Image Classification Experiments}

For image classification experiments, we testify the effectiveness of our method on several different benchmark networks on the CIFAR-10 and ImageNet datasets.

\paragraph{CIFAR-10 results.} 
We train two CNNs on CIFAR-10: one with the same architecture as in~\cite{Hinton2012}, and the other with a network-in-network architecture~\cite{Lin2014}.
Our training procedure is the same as in~\cite{Moosavi2016}. 
We still compare our Deep Defense with adversarial and Parseval training by fine-tuning from the references.
Fine-tuning hyper-parameters are summarized in the supplementary materials.
Likewise, we cut the learning rate by 2$\times$ for the last 10 epochs.

Quantitative comparison results can be found in Table~\ref{tab:rhoacc}, in which the two chosen CNNs are referred to as ``ConvNet'' and ``NIN'', respectively. 
Obviously, our Deep Defense outperforms the other defense methods considerably in all test cases.
When compared with the reference models, our regularized models achieve higher test-set accuracies on benign examples and gain absolute error decreases of 26.15\% and 16.44\% under the FGS attack.
For the DeepFool attack which might be stronger, our method gains 2.1$\times$ and 1.3$\times$ better robustness on the two networks.

\paragraph{ImageNet results.} As a challenging classification dataset, ImageNet consists of millions of high-resolution images~\cite{Russakovsky2015}.
To verify the efficacy and scalability of our method, we collect well-trained AlexNet~\cite{Krizhevsky2012} and ResNet-18~\cite{He2016} model from the Caffe and PyTorch model zoo respectively, fine-tune them on the ILSVRC-2012 training set using our Deep Defense and test it on the validation set.
After only 10 epochs of fine-tuning for AlexNet and 1 epoch for ResNet, we achieve roughly $1.5\times$ improved robustness to the DeepFool attack on both architectures, along with a slightly increased benign-set accuracy, highlighting the effectiveness of our method.

\section{Conclusion}\label{sec:con}

In this paper, we investigate the vulnerability of DNNs to adversarial examples and propose a novel method to address it, by incorporating an adversarial perturbation-based regularization into the classification objective.
This shall endow DNNs with an ability of directly learning from attacks and precisely resisting them.
We consider the joint optimization problem as learning a recursive-flavoured network to solve it efficiently.
Extensive experiments on MNIST, CIFAR-10 and ImageNet have shown the effectiveness of our method.
In particular, when combined with the FGS-based adversarial learning, our method achieves even better results on various benchmarks.
Future works shall include explorations on resisting black-box attacks and attacks in the physical world.

\subsubsection*{Acknowledgments}
This work is supported by NSFC (Grant No. 61876095, No. 61751308 and  No.61473167) and Beijing Natural Science Foundation (Grant No. L172037).


{\small
 \bibliographystyle{plain}
 \bibliography{ref}

\begin{thebibliography}{10}

\bibitem{Alemi2017}
Alexander~A Alemi, Ian Fischer, Joshua~V Dillon, and Kevin Murphy.
\newblock Deep variational information bottleneck.
\newblock In {\em ICLR}, 2017.

\bibitem{Buckman2018}
Jacob Buckman, Aurko Roy, Colin Raffel, and Ian Goodfellow.
\newblock Thermometer encoding: One hot way to resist adversarial examples.
\newblock In {\em ICLR}, 2018.

\bibitem{Carlini2017}
Nicholas Carlini and David Wagner.
\newblock Adversarial examples are not easily detected: Bypassing ten detection
  methods.
\newblock In {\em ACM Workshop on Artificial Intelligence and Security}, 2017.

\bibitem{CW2017}
Nicholas Carlini and David Wagner.
\newblock Towards evaluating the robustness of neural networks.
\newblock In {\em IEEE Symposium on Security and Privacy (SP)}, 2017.

\bibitem{Cisse2017}
Moustapha Cisse, Piotr Bojanowski, Edouard Grave, Yann Dauphin, and Nicolas
  Usunier.
\newblock Parseval networks: Improving robustness to adversarial examples.
\newblock In {\em ICML}, 2017.

\bibitem{Dhillon2018}
Guneet~S Dhillon, Kamyar Azizzadenesheli, Zachary~C Lipton, Jeremy Bernstein,
  Jean Kossaifi, Aran Khanna, and Anima Anandkumar.
\newblock Stochastic activation pruning for robust adversarial defense.
\newblock In {\em ICLR}, 2018.

\bibitem{Fawzi2016}
Alhussein Fawzi, Seyed-Mohsen Moosavi-Dezfooli, and Pascal Frossard.
\newblock Robustness of classifiers: from adversarial to random noise.
\newblock In {\em NIPS}, 2016.

\bibitem{Goodfellow2015}
Ian~J Goodfellow, Jonathon Shlens, and Christian Szegedy.
\newblock Explaining and harnessing adversarial examples.
\newblock In {\em ICLR}, 2015.

\bibitem{Gu2015}
Shixiang Gu and Luca Rigazio.
\newblock Towards deep neural network architectures robust to adversarial
  examples.
\newblock In {\em ICLR Workshop}, 2015.

\bibitem{Guo2016}
Yiwen Guo, Anbang Yao, and Yurong Chen.
\newblock Dynamic network surgery for efficient dnns.
\newblock In {\em NIPS}, 2016.

\bibitem{Han2015}
Song Han, Jeff Pool, John Tran, and William Dally.
\newblock Learning both weights and connections for efficient neural network.
\newblock In {\em NIPS}, 2015.

\bibitem{He2015}
Kaiming He, Xiangyu Zhang, Shaoqing Ren, and Jian Sun.
\newblock Delving deep into rectifiers: Surpassing human-level performance on
  imagenet classification.
\newblock In {\em ICCV}, 2015.

\bibitem{He2016}
Kaiming He, Xiangyu Zhang, Shaoqing Ren, and Jian Sun.
\newblock Deep residual learning for image recognition.
\newblock In {\em CVPR}, 2016.

\bibitem{Hein2017}
Matthias Hein and Maksym Andriushchenko.
\newblock Formal guarantees on the robustness of a classifier against
  adversarial manipulation.
\newblock In {\em NIPS}, 2017.

\bibitem{Hinton2012}
Geoffrey~E Hinton, Nitish Srivastava, Alex Krizhevsky, Ilya Sutskever, and
  Ruslan~R Salakhutdinov.
\newblock Improving neural networks by preventing co-adaptation of feature
  detectors.
\newblock {\em arXiv preprint arXiv:1207.0580}, 2012.

\bibitem{Ioffe2015}
Sergey Ioffe and Christian Szegedy.
\newblock Batch normalization: Accelerating deep network training by reducing
  internal covariate shift.
\newblock In {\em ICML}, 2015.

\bibitem{Krizhevsky2012}
Alex Krizhevsky, Ilya Sutskever, and Geoffrey~E Hinton.
\newblock Imagenet classification with deep convolutional neural networks.
\newblock In {\em NIPS}, 2012.

\bibitem{Kurakin2017}
Alexey Kurakin, Ian Goodfellow, and Samy Bengio.
\newblock Adversarial machine learning at scale.
\newblock In {\em ICLR}, 2017.

\bibitem{Lecun1999}
Yann LeCun, Patrick Haffner, L{\'e}on Bottou, and Yoshua Bengio.
\newblock Object recognition with gradient-based learning.
\newblock {\em Shape, contour and grouping in computer vision}, pages 823--823,
  1999.

\bibitem{Lin2014}
Min Lin, Qiang Chen, and Shuicheng Yan.
\newblock Network in network.
\newblock In {\em ICLR}, 2014.

\bibitem{Lu2017}
Jiajun Lu, Theerasit Issaranon, and David Forsyth.
\newblock Safetynet: Detecting and rejecting adversarial examples robustly.
\newblock In {\em ICCV}, 2017.

\bibitem{Madry2018}
Aleksander Madry, Aleksandar Makelov, Ludwig Schmidt, Dimitris Tsipras, and
  Adrian Vladu.
\newblock Towards deep learning models resistant to adversarial attacks.
\newblock In {\em ICLR}, 2018.

\bibitem{Metzen2017}
Jan~Hendrik Metzen, Tim Genewein, Volker Fischer, and Bastian Bischoff.
\newblock On detecting adversarial perturbations.
\newblock In {\em ICLR}, 2017.

\bibitem{Miyato2017}
Takeru Miyato, Shin-ichi Maeda, Masanori Koyama, and Shin Ishii.
\newblock Virtual adversarial training: a regularization method for supervised
  and semi-supervised learning.
\newblock {\em arXiv preprint arXiv:1704.03976}, 2017.

\bibitem{Moosavi2017}
Seyed-Mohsen Moosavi-Dezfooli, Alhussein Fawzi, Omar Fawzi, and Pascal
  Frossard.
\newblock Universal adversarial perturbations.
\newblock In {\em CVPR}, 2017.

\bibitem{Moosavi2016}
Seyed-Mohsen Moosavi-Dezfooli, Alhussein Fawzi, and Pascal Frossard.
\newblock Deep{F}ool: a simple and accurate method to fool deep neural
  networks.
\newblock In {\em CVPR}, 2016.

\bibitem{Noh2015}
Hyeonwoo Noh, Seunghoon Hong, and Bohyung Han.
\newblock Learning deconvolution network for semantic segmentation.
\newblock In {\em ICCV}, 2015.

\bibitem{Papernot2017}
Nicolas Papernot, Patrick McDaniel, Ian Goodfellow, Somesh Jha, Z~Berkay Celik,
  and Ananthram Swami.
\newblock Practical black-box attacks against machine learning.
\newblock In {\em Asia Conference on Computer and Communications Security},
  2017.

\bibitem{Papernot2018}
Nicolas Papernot, Patrick McDaniel, Arunesh Sinha, and Michael Wellman.
\newblock Towards the science of security and privacy in machine learning.
\newblock In {\em IEEE European Symposium on Security and Privacy}, 2018.

\bibitem{Papernot2016}
Nicolas Papernot, Patrick McDaniel, Xi~Wu, Somesh Jha, and Ananthram Swami.
\newblock Distillation as a defense to adversarial perturbations against deep
  neural networks.
\newblock In {\em IEEE Symposium on Security and Privacy (SP)}, 2016.

\bibitem{Ross2018}
Andrew~Slavin Ross and Finale Doshi-Velez.
\newblock Improving the adversarial robustness and interpretability of deep
  neural networks by regularizing their input gradients.
\newblock In {\em AAAI}, 2018.

\bibitem{Russakovsky2015}
Olga Russakovsky, Jia Deng, Hao Su, Jonathan Krause, Sanjeev Satheesh, Sean Ma,
  Zhiheng Huang, Andrej Karpathy, Aditya Khosla, Michael Bernstein,
  Alexander~C. Berg, and Li~Fei-Fei.
\newblock Imagenet large scale visual recognition challenge.
\newblock {\em IJCV}, 2015.

\bibitem{Sokolic2017}
Jure Sokolic, Raja Giryes, Guillermo Sapiro, and Miguel~RD Rodrigues.
\newblock Robust large margin deep neural networks.
\newblock {\em IEEE Transactions on Signal Processing}, 2017.

\bibitem{Szegedy2014}
Christian Szegedy, Wojciech Zaremba, Ilya Sutskever, Joan Bruna, Dumitru Erhan,
  Ian Goodfellow, and Rob Fergus.
\newblock Intriguing properties of neural networks.
\newblock In {\em ICLR}, 2014.

\bibitem{Xie2018}
Cihang Xie, Jianyu Wang, Zhishuai Zhang, Zhou Ren, and Alan Yuille.
\newblock Mitigating adversarial effects through randomization.
\newblock In {\em ICLR}, 2018.

\bibitem{Xie2017}
Cihang Xie, Jianyu Wang, Zhishuai Zhang, Yuyin Zhou, Lingxi Xie, and Alan
  Yuille.
\newblock Adversarial examples for semantic segmentation and object detection.
\newblock In {\em ICCV}, 2017.

\bibitem{Xu2012}
Huan Xu and Shie Mannor.
\newblock Robustness and generalization.
\newblock {\em Machine learning}, 86(3):391--423, 2012.

\bibitem{Zeiler2014}
Matthew~D Zeiler and Rob Fergus.
\newblock Visualizing and understanding convolutional networks.
\newblock In {\em ECCV}, 2014.

\end{thebibliography}
}

\newpage


\section*{Supplementary material}

# Supplementary Material for Deep Defense: Training DNNs with Improved Adversarial Robustness

**Ziang Yan**[1*]   **Yiwen Guo**[2,1*]   **Changshui Zhang**[1]
[1]Institute for Artificial Intelligence, Tsinghua University (THUAI),
State Key Lab of Intelligent Technologies and Systems,
Beijing National Research Center for Information Science and Technology (BNRist),
Department of Automation,Tsinghua University, Beijing, China
[2] Intel Labs China
yza18@mails.tsinghua.edu.cn   yiwen.guo@intel.com   zcs@mail.tsinghua.edu.cn

## 1   More Evaluation Metrics and Attacks

Table 2: Test performance of different methods in the sense of: $l_2$ under DeepFool, $\rho_\infty$ under FGS, $l_2$ under the C&W attack, the prediction accuracy on PGD adversarial examples and the PASS score.

| Dataset | Network | Method | $l_2$ (DeepFool) | $\rho_\infty$ (FGS) | $l_2$ (C&W) | Acc. (PGD) | PASS |
|---|---|---|---|---|---|---|---|
| MNIST | MLP | Reference | 0.81 | $5.40 \times 10^{-2}$ | 0.84 | 1.19% | 0.8534 |
| | | Par. Train | 0.80 | $5.78 \times 10^{-2}$ | 0.83 | 1.18% | 0.8542 |
| | | Adv. Train I | 1.17 | $9.46 \times 10^{-2}$ | 1.17 | 4.11% | 0.8280 |
| | | Deep Defense | **1.64** | $\mathbf{1.53 \times 10^{-1}}$ | **1.58** | **33.18%** | **0.8181** |
| | LeNet | Reference | 1.48 | $1.29 \times 10^{-1}$ | 1.40 | 26.17% | 0.9074 |
| | | Par. Train | 1.50 | $1.50 \times 10^{-1}$ | 1.58 | 23.06% | 0.8981 |
| | | Adv. Train I | 1.90 | $2.05 \times 10^{-1}$ | 1.71 | 50.67% | 0.8810 |
| | | Deep Defense | **2.05** | $\mathbf{2.36 \times 10^{-1}}$ | **1.84** | **64.54%** | **0.8760** |
| CIFAR-10 | ConvNet | Reference | 0.18 | $5.27 \times 10^{-3}$ | 0.29 | 21.34% | - |
| | | Par. Train | 0.24 | $8.02 \times 10^{-3}$ | 0.33 | 27.91% | - |
| | | Adv. Train I | 0.21 | $6.37 \times 10^{-3}$ | 0.31 | 27.08% | - |
| | | Deep Defense | **0.36** | $\mathbf{1.58 \times 10^{-2}}$ | **0.47** | **45.05%** | - |
| | NIN | Reference | 0.30 | $1.05 \times 10^{-2}$ | 0.41 | 34.41% | - |
| | | Par. Train | 0.31 | $1.07 \times 10^{-2}$ | 0.41 | 36.59% | - |
| | | Adv. Train I | 0.37 | $1.76 \times 10^{-2}$ | 0.48 | 45.51% | - |
| | | Deep Defense | **0.40** | $\mathbf{2.15 \times 10^{-2}}$ | **0.50** | **51.07%** | - |
| ImageNet | AlexNet | Reference | 0.29 | $5.46 \times 10^{-4}$ | - | - | - |
| | | Deep Defense | **0.45** | $\mathbf{8.70 \times 10^{-4}}$ | - | - | - |
| | ResNet | Reference | 0.69 | $6.96 \times 10^{-4}$ | - | - | - |
| | | Deep Defense | **1.03** | $\mathbf{1.08 \times 10^{-3}}$ | - | - | - |

In this paper, we leave the optimal choice of evaluation metric an open question and simply choose some popular ones following previous works. Here in the supplementary material we try to test as many as possible to verify the effectiveness of our method extensively.

In the main body of our paper, we utilize the normalized $l_2$ norm of required adversarial perturbations to evaluate the robustness of different models, as suggested in the DeepFool paper [6]. We notice that in some papers, an unnormalized norm is used instead, which means

$$l_2 := \frac{1}{|\mathcal{D}|} \sum_{k \in \mathcal{D}} \|\Delta_{\mathbf{x}_k}\|_2 \tag{1}$$

---

[*]The first two authors contributed equally to this work.

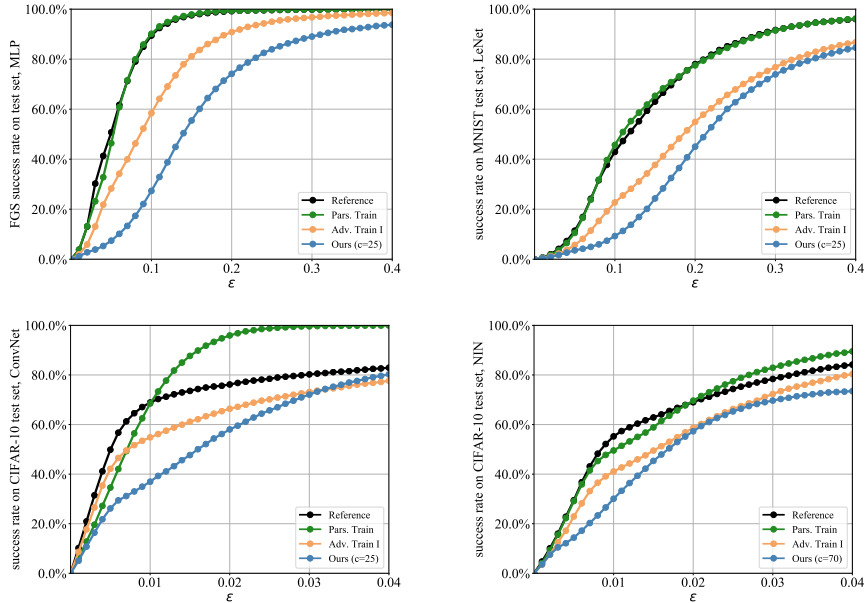

Figure 5: Comparison of different defense methods under the FGS attack. For each network, we report the success rate of FGS with varying $\epsilon$. **Lower is better**. Best viewed in color.

can also be calculated as a metric (see the fourth column of Table 2). In addition, we further evaluate the robustness of different models under the C&W's $l_2$ attack [1], using the official CleverHans [7] implementation. The (unnormalized) $l_2$ values under the C&W's $l_2$ attack are reported in the sixth column of Table 2. Using different reference models trained with different initializations lead to very similar results in our experiments, so we simply omit such variance (e.g., for $l_2$, it is less than 0.003).

Also, when the FGS attack is adopted, the robustness can be evaluated by replacing the $l_2$ norm with an $l_\infty$ norm in the definition of $\rho$ as the FGS attack is usually considered as an $l_\infty$ norm-based (or max-norm based) perturbation method, and get

$$\rho_\infty := \frac{1}{|\mathcal{D}|} \sum_{k \in \mathcal{D}} \frac{\|\Delta_{\mathbf{x}_k}\|_\infty}{\|\mathbf{x}_k\|_\infty}. \qquad (2)$$

in the fifth column of Table 2. Higher $l_2$ and $\rho_\infty$ indicate better robustness to the $l_2$ and $l_\infty$ attacks, respectively. Recall that, to establish a benchmark, we adjust $\epsilon$ such that 50% of the image samples are misclassified by well-trained models, as introduced in the main body of our paper. Here we further compare the FGS success rates with respect to varying $\epsilon$ on different models in Figure 5.

As an additional $l_\infty$ attack, the PGD-based method [4] is also tested here. We set $\epsilon = 0.1$ for MNIST, $\epsilon = 0.01$ for CIFAR-10, and compare prediction accuracies on adversarial examples in the seventh column of Table 2. It can be seen that the superiority of our method holds on various baseline networks. Recently, Rozsa et.al. [8] propose a psychometric perceptual adversarial similarity score named *PASS*, which seems consistent with human perception. The lower such score is, the better defensive performance the model gets. We calculate it using an official implementation provided by the authors and report some results in the last column of Table 2.

## 2 Comparison with Adv. Train II

As introduced in the main body of our paper, various forms of adversarial training have been adopted in previous works [10, 2, 6, 3, 5, 4]. Here we test Goodfellow et al.'s adversarial training (abbreviated as "Adv. Train II"). In addition, we also try combining it with our Deep Defense by simply adding the cross entropy loss corresponding to FGS adversarial examples to the training objective of our method. The performance of Adv. Train II, our Deep Defense and our combined method are compared in Figure 6. For each network, we report the $\rho_2$ values under DeepFool in the left column and success rate of FGS with varying $\epsilon$ in the right column.

For our Deep Defense, we fix $\lambda$ and $d$ and vary only $c$ in the figure, while for the combined method, we further fix $c$ and vary only $\epsilon$, as for Adv. Train II. In the right column, we select winning Adv.

2

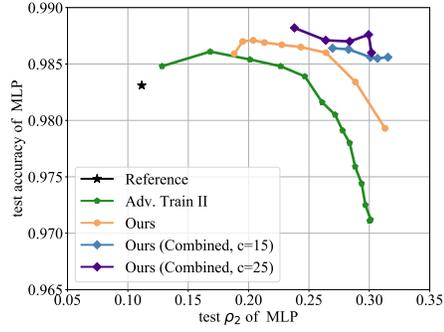
(a)

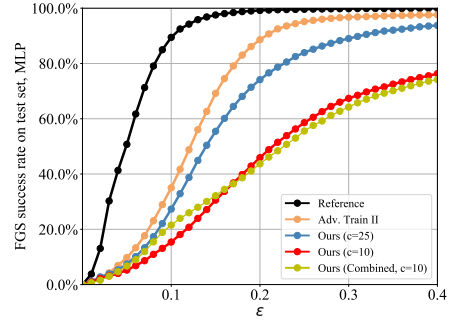
(b)

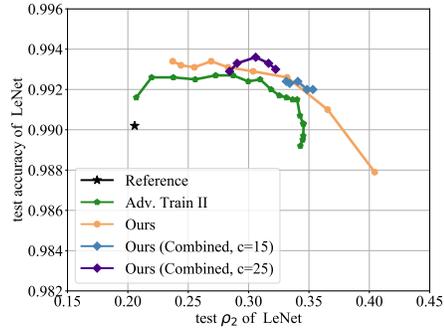
(c)

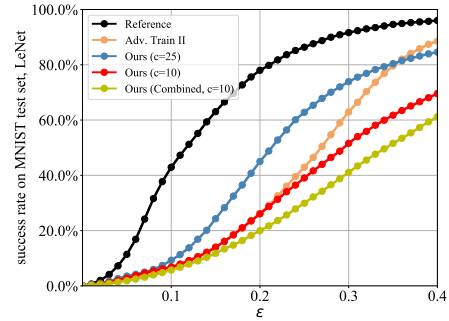
(d)

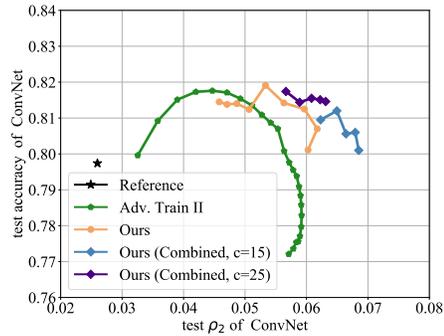
(e)

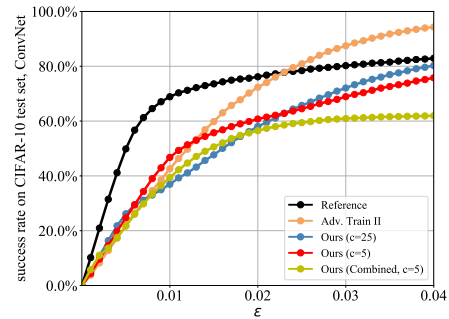
(f)

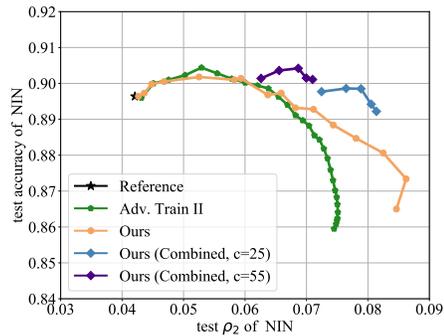
(g)

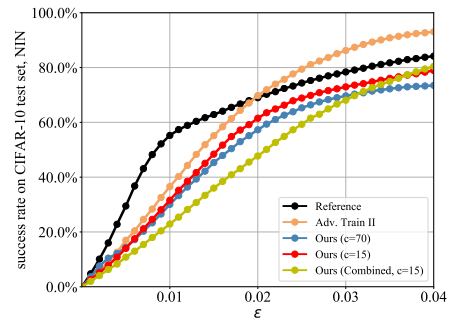
(h)

Figure 6: Comparison with Adv. Train II on both MNIST and CIFAR-10 datasets. For each network, we report the $\rho_2$ values with DeepFool in the left column (**upper right is better**) and the success rate of FGS with varying $\epsilon$ in the right column (**lower is better**). Best viewed in color.

3

Train II models (under the FGS attack) from those tested in the left. Obviously, we see that our Deep Defense outperforms Adv. Train II as well in most cases. Moreover, by combining them, we gain even better robustness and benign-set accuracies, which verifies our previous claim of orthogonality.

## 3 MNIST Visualization Results

Quantitative results in our paper demonstrate that an adversary has to generate larger perturbations to successfully attack our regularized models. Intuitively, this implies that the required perturbations should be perceptually more obvious. Here we provide visualization results in Figure 7. Given a clean image from the test set (as illustrated in Figure 7(a)), the generated DeepFool adversarial examples for successfully fooling different models are shown in Figure 7(b)-7(e). Obviously, our method yields more robust models in comparison with the others, by making the adversarial examples closely resembling real "8" and "6" images. More interestingly, our regularized LeNet model predicts all examples in Figure 7(a)-7(d) correctly as "0". For the lower adversarial example in Figure 7(e), it makes the correct prediction "0" with a probability of 0.30 and the incorrect one (i.e., "6") with a probability of 0.69.

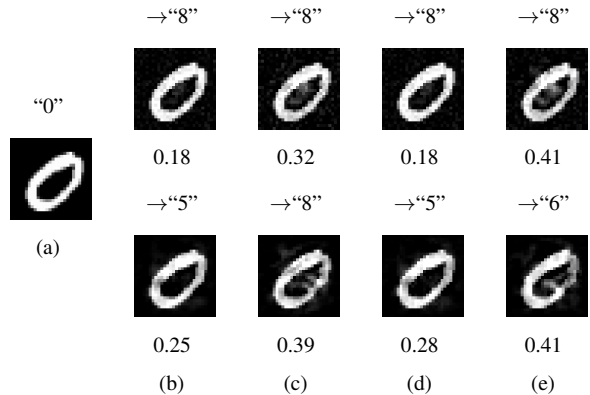

Figure 7: An image ($\mathbf{x}_k$) labelled "0" from the MNIST test set with DeepFool examples generated to fool different models including: (b) the references, (c)-(e): fine-tuned models with Adv. Train I, Parseval training and our Deep Defense method. Arrows above the pictures indicate which class the examples are "misclassified" to and the numbers below indicate values of $\|\Delta_{\mathbf{x}_k}\|_2/\|\mathbf{x}_k\|_2$. Upper images are generated for MLP models and lower images are generated for LeNet models.

## 4 CIFAR-10 Convergence Curves

Convergence curves on CIFAR-10 of different methods are provided in Figure 8.

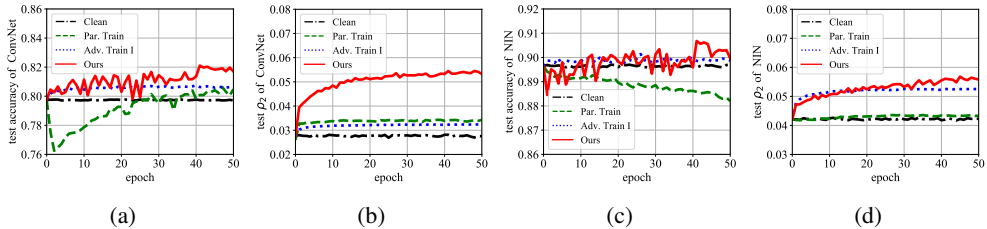

Figure 8: Convergence curves on CIFAR-10: (a)-(b) test accuracy and $\rho_2$ of ConvNet, and (c)-(d) test accuracy and $\rho_2$ of NIN. "Clean" indicates fine-tuning on benign examples. Best viewed in color.

## 5 ImageNet Results

Our method yields models with substantially improved robustness and no accuracy loss is observed on benign test sets, even on ImageNet. Though also enhance models, Parseval and adversarial training

4

seem difficult to achieve good trade-offs between robustness and accuracy in our experiments on ImageNet. On AlexNet, we were unable to find a suitable $\beta$ such that the fine-tuned model shows reasonably high accuracy ($> 56\%$) and significantly improved robustness simultaneously for Parseval training. This phenomenon can possibly be caused by insufficient hyper-parameter search. For Adv. Train I and II, we observed a decrease of inference accuracy on benign examples when the fine-tuning process starts, and after 10 epochs the accuracy is still unsatisfactory. However, Kurakin et.al. [3] have produced an Inception v3 model [9] using 50 machines after $150k$ iterations (i.e.roughly 187 epochs) of training and obtain only slightly degraded accuracy, so we guess more training epochs and sophisticated mixture of clean and adversarial examples are required.

## 6 Network Architectures and Hyper-parameters

Some hyper-parameters for our fine-tuning are summarized in Table 3. Other hyper-parameters like momentum and weight decay are kept as the same as training the reference models (i.e., momentum: 0.9, and weight decay: 0.0005). Table 4 shows the architecture of networks used in our MNIST and CIFAR-10 experiments. For AlexNet and ResNet experiments, we directly use the reference models from the Caffe and PyTorch model zoos.

Table 3: Some hyper-parameters in the fine-tuning process.

| Dataset | Batch Size | #Epoch | Base Learning Rate |
|---|---|---|---|
| MNIST | 100 | 5 | $5 \times 10^{-4}$ |
| CIFAR-10 | 100 | 50 | $5 \times 10^{-4}$ |
| ImageNet | 256 | 10 | $1 \times 10^{-4}$ |

Table 4: Network architectures adopted in MNIST and CIFAR-10 experiments. We use Conv-[kernel width]-[output channel number], FC-[output channel number], MaxPool-[kernel width], AvgPool-[kernel width] to denote parameters of convolutional layers, fully-connected layers, max pooling layers and average pooling layers, respectively.

| MNIST | | CIFAR-10 | |
|---|---|---|---|
| MLP | LeNet | ConvNet | NIN |
| Input (28×28) | | Input (32×32) | |
| FC-500 | Conv-5-20 | Conv-5-32 | Conv-5-192 |
| ReLU | MaxPool-2 | MaxPool-2 | ReLU |
| FC-150 | Conv-5-50 | ReLU | Conv-1-160 |
| ReLU | MaxPool-2 | Conv-5-32 | ReLU |
| FC-10 | FC-500 | ReLU | Conv-1-96 |
| | ReLU | AvgPool-2 | ReLU |
| | FC-10 | Conv-5-64 | MaxPool-2 |
| | | ReLU | Conv-5-192 |
| | | AvgPool-2 | ReLU |
| | | Conv-4-64 | Conv-1-192 |
| | | ReLU | ReLU |
| | | Conv-1-10 | Conv-1-192 |
| | | | ReLU |
| | | | AvgPool-2 |
| | | | Conv-3-192 |
| | | | ReLU |
| | | | Conv-1-192 |
| | | | ReLU |
| | | | Conv-1-10 |
| | | | AvgPool-8 |

# 7  Note on Max-unpooling Layers of the Reverse Network

In the main body of our paper, we mimic the DeepFool attack calculation using a neural network. In order to do this, the forward process of the "reverse" network should generate an exact output as the backward process of the original classification network. As discussed in the main paper, feasible "reverse" layers can always be made available when building the reverse network.

Special attention should be paid when reversing max-pooling layers. In many modern DL frameworks (including TF, PyTorch and Keras), the forward process of a max-unpooling layer is not strictly equal to the backward process of a max pooling layer, if the stride is smaller than the pooling window size. In the max pooling operation, if more than one overlapped sliding windows select the same element from feature maps simultaneously, the derivatives from later feature maps should be summed up in the backward calculation. However, many DL frameworks just select one of the overlapped windows and ignore the others in the forward process of a max unpooling, which is slightly different. Such difference could accumulate layer-by-layer and the final perturbation can be very different from the original DeepFool, especially for deep networks. We release a patch to fix this along with our source code at `https://github.com/ZiangYan/deepdefense.pytorch`.

## References

[1] Nicholas Carlini and David Wagner. Towards evaluating the robustness of neural networks. In *IEEE Symposium on Security and Privacy (SP)*, 2017.

[2] Ian J Goodfellow, Jonathon Shlens, and Christian Szegedy. Explaining and harnessing adversarial examples. In *ICLR*, 2015.

[3] Alexey Kurakin, Ian Goodfellow, and Samy Bengio. Adversarial machine learning at scale. In *ICLR*, 2017.

[4] Aleksander Madry, Aleksandar Makelov, Ludwig Schmidt, Dimitris Tsipras, and Adrian Vladu. Towards deep learning models resistant to adversarial attacks. In *ICLR*, 2018.

[5] Takeru Miyato, Shin-ichi Maeda, Masanori Koyama, and Shin Ishii. Virtual adversarial training: a regularization method for supervised and semi-supervised learning. *arXiv preprint arXiv:1704.03976*, 2017.

[6] Seyed-Mohsen Moosavi-Dezfooli, Alhussein Fawzi, and Pascal Frossard. DeepFool: a simple and accurate method to fool deep neural networks. In *CVPR*, 2016.

[7] Nicolas Papernot, Nicholas Carlini, Ian Goodfellow, Reuben Feinman, Fartash Faghri, Alexander Matyasko, Karen Hambardzumyan, Yi-Lin Juang, Alexey Kurakin, Ryan Sheatsley, et al. cleverhans v2.0.0: an adversarial machine learning library. *arXiv preprint arXiv:1610.00768*, 2017.

[8] Andras Rozsa, Ethan M Rudd, and Terrance E Boult. Adversarial diversity and hard positive generation. In *CVPR Workshop*, 2016.

[9] Christian Szegedy, Wei Liu, Yangqing Jia, Pierre Sermanet, Scott Reed, Dragomir Anguelov, Dumitru Erhan, Vincent Vanhoucke, and Andrew Rabinovich. Going deeper with convolutions. In *CVPR*, pages 1–9, 2015.

[10] Christian Szegedy, Wojciech Zaremba, Ilya Sutskever, Joan Bruna, Dumitru Erhan, Ian Goodfellow, and Rob Fergus. Intriguing properties of neural networks. In *ICLR*, 2014.


\end{document}